\documentclass{article}
\usepackage{ijcai23}

\usepackage{times}
\usepackage{soul}
\usepackage{url}
\usepackage[hidelinks]{hyperref}
\usepackage[utf8]{inputenc}
\usepackage[small]{caption}
\usepackage{graphicx}
\usepackage{amsmath}
\usepackage{amsthm}
\usepackage{booktabs}
\usepackage{algorithm}
\usepackage{algorithmic}
\usepackage[switch]{lineno}
\usepackage{amsmath,amsfonts}
\usepackage{array}
\usepackage{textcomp}
\usepackage{stfloats}
\usepackage{verbatim}
\usepackage{graphicx}
\usepackage{boldline}
\usepackage{multirow}
\usepackage{graphicx}
\usepackage{diagbox}
\usepackage{amssymb}
\usepackage{paralist,bm}
\usepackage{makecell}
\usepackage{subfigure}
\usepackage{xcolor}
\usepackage[hidelinks]{hyperref}
\renewcommand{\vec}[1]{\bm{#1}}

\def\eg{\emph{e.g.~}} 
\def\ie{\emph{i.e.~}}

\def\wrt{w.r.t.~} 
\def\etal{\emph{et al}.~}

\newenvironment{packed_itemize}{
	\begin{itemize}
		\setlength{\itemsep}{1pt}
		\setlength{\parskip}{0pt}
		\setlength{\parsep}{0pt}
	}{\end{itemize}}

\title{Manifold-Aware Self-Training for Unsupervised Domain Adaptation on Regressing 6D Object Pose}

\author{
Yichen Zhang$^1$
\and
Jiehong Lin$^1$\and
Ke Chen$^{1, 2, }$\footnote{Corresponding authors.}\and
Zelin Xu$^1$\and
Yaowei Wang$^2$\And
Kui Jia$^{1, 2,*}$
\affiliations
$^1$South China University of Technology\\
$^2$Peng Cheng Laboratory
\emails
\{eezyc, lin.jiehong, eexuzelin\}@mail.scut.edu.cn,
\{chenk, kuijia\}@scut.edu.cn,
wangyw@pcl.ac.cn
}

\begin{document}
\maketitle

\begin{abstract}
Domain gap between synthetic and real data in visual regression (\eg 6D pose estimation) is bridged in this paper via global feature alignment and local refinement on the coarse classification of discretized anchor classes in target space, which imposes a piece-wise target manifold regularization into domain-invariant representation learning. Specifically, our method incorporates an explicit self-supervised manifold regularization, revealing consistent cumulative target dependency across domains, to a self-training scheme (\eg the popular Self-Paced Self-Training) to encourage more discriminative transferable representations of regression tasks. Moreover, learning unified implicit neural functions to estimate relative direction and distance of targets to their nearest class bins aims to refine target classification predictions, which can gain robust performance against inconsistent feature scaling sensitive to UDA regressors. Experiment results on three public benchmarks of the challenging 6D pose estimation task can verify the effectiveness of our method, consistently achieving superior performance to the state-of-the-art for UDA on 6D pose estimation. Code is available at \url{https://github.com/Gorilla-Lab-SCUT/MAST}.

\end{abstract}

\section{Introduction}
\label{sec:intro}

\begin{figure}[t]
    \centering
    \includegraphics[width=0.95\columnwidth]{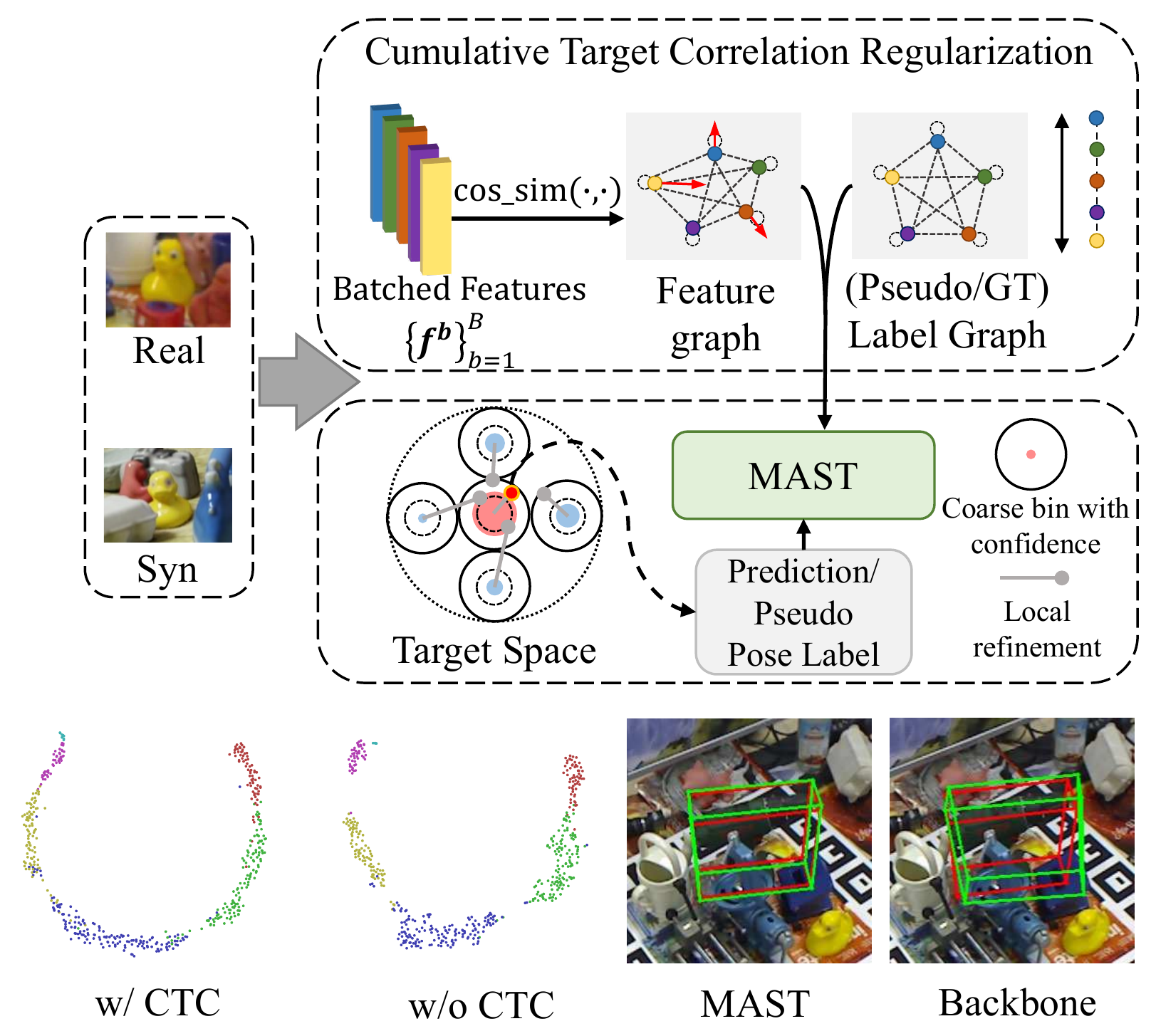}
    \caption{Illustration of the proposed Manifold-Aware Self Training (MAST) for UDA on 6D pose estimation. 
    Top: a novel cumulative target correlation (CTC) regularization on representation learning. 
    Middle: one bin with the highest confidence (\ie highlighted with the largest red dish) with further local refinement (\ie the gray lines) are adopted to generate pseudo pose labels in our MAST. 
    Bottom: the t-SNE visualization of feature distribution w/ and w/o the proposed CTC, which can verify the effectiveness of introduction of target manifolds with a more smooth distribution; and comparison with our MAST and its backbone with an example from the Occluded LineMOD dataset is given.}
    \label{fig:intro}
\end{figure}

The problems of visual regression such as estimation of 6D pose of object instances (\ie their orientation and translation with respect to the camera optical center) and configuration of human body parts given RGB images are widely encountered in numerous fields such as robotics \cite{collet2011moped,he2020pvn3d}, augmented reality \cite{marchand2015pose} and autonomous driving \cite{geiger2012we,chen2017multi}, which can be typically addressed by learning a single or multi-output regression mapping on deep representations of visual observations. 
Recent regression algorithms have gained remarkable success to handle with inconsistent lighting conditions and heavy occlusions in-between foreground and contextual objects in uncontrolled and cluttered environment, owing to recent development of representation learning of visual regression, such as introduction of self-supervised regularization \cite{lin2022category} and powerful network architectures \cite{he2021ffb6d}.

In those regression tasks, visual observations, \ie RGB images, can be easily acquired in practice or directly collected from the Internet, but it is laborious or even unfeasible for manual noise-free annotation with continuous targets.  
As a result, the size of real training data with precise labels is typically limited and less scalable, \eg \textit{eggbox} and \textit{holepuncher} training samples in the LineMOD \cite{hinterstoisser2012model} for 6D pose estimation, which increases the difficulty of learning good representations.
The synthesis of images can be a powerful solution to cope with data sparsity, which can be gained via photorealistic rendering \cite{hodavn2019photorealistic} with CAD models.
However, domain discrepancy between synthetic and real data, \eg appearance difference between CAD models and real objects, scene illumination, and systematic imaging noises, can lead to collapse of regression performance, which encourages the practical setting of unsupervised domain adaptation on visual regression (UDAVR), \ie samples in the source and target domains cannot satisfy the i.i.d. condition. 

Different from the widely-investigated problem of unsupervised domain adaptation on visual classification (UDAVC) \cite{gopalan2011domain,zou2018unsupervised,zou2021geometry}, only a few works \cite{chen2021representation,lee2022uda} have explored the vital factors of representation learning of visual regression that different from classification in the context of UDA.
\cite{chen2021representation} revealed and exploited the sensitivity of feature scaling on domain adaptation regression performance to regularize representation learning, which can achieve promising results to bridge domain gap. 
We argue that \emph{cumulative dependent nature} and \emph{piece-wise  manifolds} in target space are two key factors of UDA regression yet missing in the existing algorithms.
{To this end, this paper proposes a \textbf{M}anifold-\textbf{A}ware \textbf{S}elf-\textbf{T}raining (MAST) scheme to decompose the problem of learning a domain-invariant regression mapping into a combination of a feature-scaling-robust globally coarse classification of discretized target anchors via self-training based feature alignment and a locally regression-based refinement less sensitive to inconsistent feature scale, as shown in Figure \ref{fig:intro}.}

For exploiting the cumulative dependent nature of regression targets different from those in classification, the self-training method (\eg the self-paced self-training \cite{zou2018unsupervised}) originally designed for the UDAVC problem is now adapted to the coarse classification on discretization of continuous target space, with incorporating a novel piece-wise manifold regularization on domain-invariant representation learning, namely a self-supervised cumulative target correlation regularization. 
Intuitively, appearance ambiguities across domains in representation learning can be mitigated via leveraging consistent target correlation under certain distance metrics in target space (\eg the Euclidean distance in $R(3)$ translation space).  
{Furthermore, considering the risk of sensitivity to varying feature scaling in the UDAVR problem} \cite{chen2021representation}, learning unified local regression functions with those shared features of the classification of discretized target bins (typically having inconsistent feature scales) can achieve superior robustness against large scale variations of transferable representations.   
Extensive experiments on three popular benchmarks of the challenging UDA on 6D pose estimation can confirm the effectiveness of our MAST scheme, consistently outperforming the state-of-the-art.

The novelties of our paper are summarized as follows.
\begin{packed_itemize}
\item This paper proposes a novel and generic manifold-aware self-training scheme for unsupervised domain adaptation on visual regression, which exploits cumulative correlation and piece-wise manifolds in regression target space for domain-invariant representation learning.  
\item Technically, a novel cumulative target correlation regularization is proposed to regularize the self-training algorithm on coarse classification with latent dependency across regression targets, while local refinement can be achieved via learning implicit functions to estimate residual distance to the nearest anchor within local target manifolds in a unified algorithm.   
\item Experiment results on multiple public benchmarks of UDA on 6D pose estimation can verify consistent superior performance of our scheme to the state-of-the-art UDA pose regressors. 
\end{packed_itemize}

\section{Related Works}
\paragraph{6D Pose Estimation.}
The problem of estimating 6D pose of object instances within an RGB image (optionally with a complementary depth image) is active yet challenging in  robotics and computer vision. 
With the rise of deep learning, recent methods for predicting 6D poses can be divided into two main groups -- keypoint-based \cite{peng2019pvnet,zakharov2019dpod} and regression based \cite{xiang2017posecnn,labbe2020cosypose,wang2021gdr}.
The former relied on learning a 2D-to-3D correspondence mapping between  object keypoints in 3D space and their 2D projection on images with the Perspective-n-Point (PnP) \cite{fischler1981random}.
Such a correspondence can be achieved by either detecting a limited size of landmarks \cite{tekin2018real,peng2019pvnet} or pixel-wise voting from a heatmap \cite{park2019pix2pose,zakharov2019dpod}.  
The latter concerned on deep representation learning for direct pose regression with the point-matching loss for optimizing output pose \cite{xiang2017posecnn,labbe2020cosypose} or proposing a differentiable PnP paradigm in an end-to-end training style \cite{wang2021gdr,chen2022epro}. 
Alternatively, the problem can also be formulated into ordinal classification via discretization of $SE(3)$ space into class bins \cite{su2015render,kehl2017ssd}.
To alleviate representation ambiguities, the estimated 6D pose of objects can be further refined via either an iterative refinement with residual learning \cite{li2018deepim,manhardt2018deep} or simply the Iterative Closest Point \cite{xiang2017posecnn}, while some work introduced cross-view fusion based refinement \cite{labbe2020cosypose,li2018unified}.
Existing refinement strategies are typically employed as a post-processing step following the main module of 6D pose estimation, some of which such as \cite{li2018deepim,xu2022rnnpose} can be designed in an end-to-end learning cascade to obtain significant performance gain, but they are not designed for bridging domain gap and therefore cannot ensure good performance under the UDAVR setting. 
Alternatively, \cite{li2018unified} introduced a combined scheme of both coarse classification and local regression-based refinement simultaneously, which is similar to our MAST method.
However, the main differences lie in the introduction of the cumulative target correlation regularization in our scheme to encourage domain-invariant pose representations revealing the dependent nature of regression targets.

\paragraph{Unsupervised Domain Adaptation on Visual Regression.} Most of regression methods \cite{xu2019w,bao2022generalizing} employ annotated real data for model training, but manual annotations on real data are usually laboriously expensive or even unfeasible.
Lack of sufficient annotated real data encourages the practical setting of
Simulation-to-Reality (Sim2Real) UDAVR, \ie learning a domain-agnostic representation given annotated synthetic data as source domain and unlabeled real data as target domain.      
A simple yet effective way to narrow Sim2Real domain gap can rely on domain randomization \cite{kehl2017ssd,manhardt2018deep}, while recent success of self-supervised learning for UDAVC \cite{zou2021geometry,yue2021prototypical} inspired a number of self-supervised regressors \cite{wang2021occlusion,yang2021dsc} in the context of Regression. 
Self6D \cite{wang2020self6d} and its extension Self6D++ \cite{wang2021occlusion} leveraged a differentiable renderer to conduct self-supervised visual and geometrical alignment on visible and amodal object mask predictions.
Bao \etal \cite{bao2022generalizing} introduced a self-supervised representation learning of relative rotation estimation to adapt one gaze regressor to the target domain.
Zhang \etal \cite{zhang2021keypoint} utilized a Graph Convolutional Network  to model domain-invariant geometry structure among key-points, which is applied to guide training of the object pose estimator on real images. 
These mentioned algorithms were designed for only one specific task and cannot be directly applied to other visual regression  problems.
\cite{chen2021representation} proposed the representation subspace distance (RSD) generic to multiple UDAVR problems, but cannot perform well on the challenging task having severe representation ambiguities, \eg 6D pose estimation investigated in this paper (see Table \ref{tab:ablation_LM-LMO}).      
In contrast, the proposed MAST scheme is generic to UDAVR owing to exploiting explicit target correlation in the style of local manifolds to regularize deep representation learning agnostic to domains.

\paragraph{Self-Training.} 
Self-training methods utilize a trained model on labeled data to make predictions of unannotated data as pseudo labels \cite{lee2013pseudo} (\ie supervision signals assigned to unlabeled data), which is widely used in semi-supervised learning \cite{lee2013pseudo,sohn2020fixmatch} and UDA \cite{roy2019unsupervised}. 
\cite{sohn2020fixmatch} generated pseudo labels from weakly augmented images, which are adopted as supervision of strongly augmented variants in semi-supervised learning; similar scripts are shared with the noisy student training \cite{xie2020self}. 
\cite{chen2011co} proposed the co-training for domain adaptation that slowly adding to the training set both the target features and instances in which the current algorithm is the most confident. 
\cite{zou2018unsupervised} proposed the self-paced self-training (SPST) for unsupervised domain adaptation classification that can perform a self-paced learning \cite{tang2012shifting} with latent variable objective optimization. 
The representative SPST has inspired a number of follow-uppers such as \cite{zou2021geometry} and \cite{chen2022quasi}. 
Nevertheless, all of existing self-training algorithms were designed for classification or segmentation, while self-training for the UDAVR remains a promising yet less explored direction.

\section{Methodology}

Given a source domain $\{\mathcal{I}^i_S, \vec{y}^i_S\}_{i=1}^{N_S}$ with $N_S$ labeled samples and a target domain $\{\mathcal{I}^i_T\}_{i=1}^{N_T}$ with $N_T$ unlabeled samples, tasks of UDAVR aim at learning a domain-invariant regression mapping to a shared continuous label space $\mathcal{Y}$. In the context of our focused 6D object pose estimation, the source and target domains are often the synthetic and real-world data, respectively, while the shared label space between two domains is the whole learning space of $SE(3)$.

To deal with the problems of UDAVR introduced in Sec. \ref{sec:intro}, \eg, cumulative dependent nature and piece-wise manifolds in target space, we propose in this paper a manifold-aware self-training scheme, which decomposes the learning of $SE(3)$ space into a global classification on discretized pose anchors and a local pose refinement for feature scaling robustness, and incorporates a self-supervised manifold regularization to the self-training.

\begin{figure*}[htbp]
    \centering
    \includegraphics[width=0.9\linewidth]{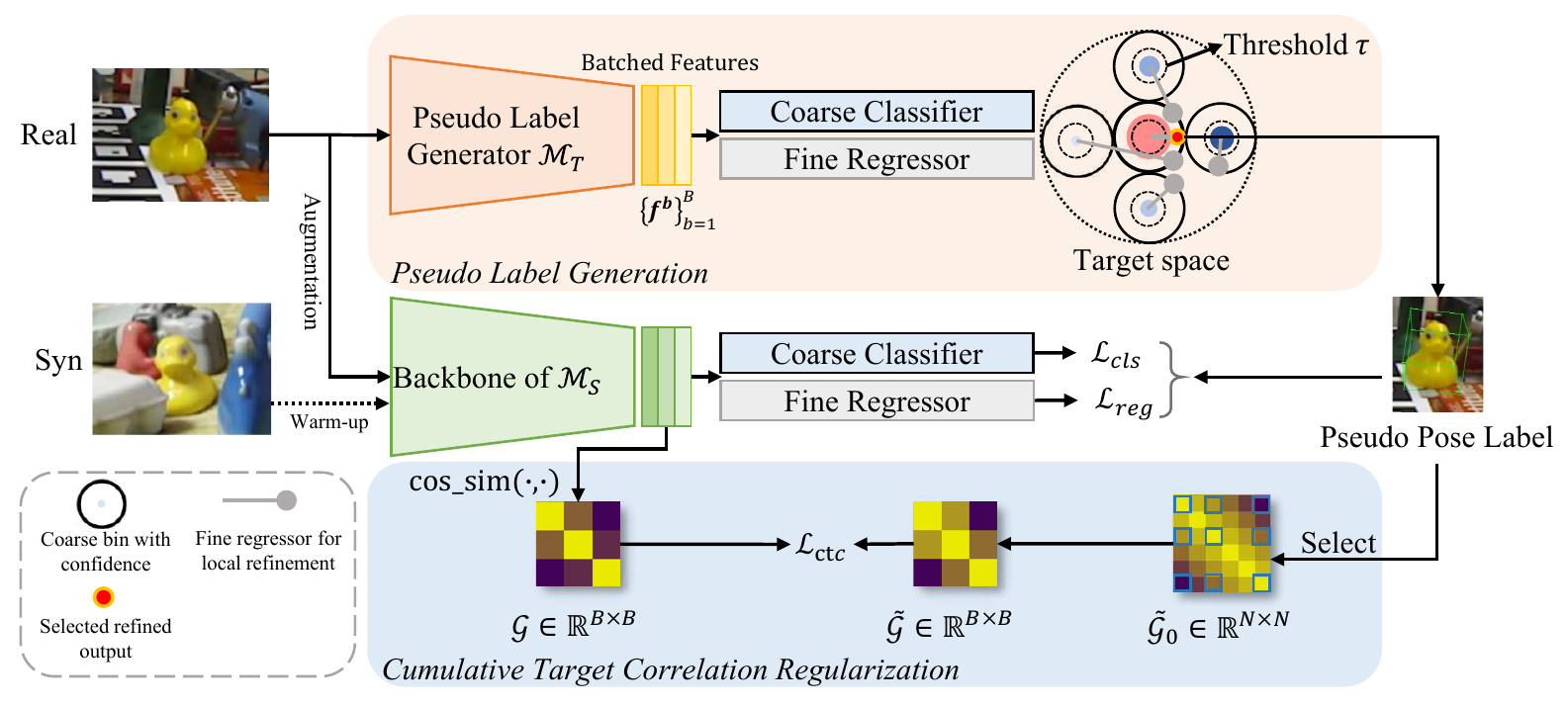}
    \caption{The pipeline of our manifold-aware self-training scheme.}
    \label{fig:model_arch}
\end{figure*}

\subsection{The Design of Network Architecture}
\label{subsec:network}

Given a batch of $B$ object-centric RGB images $\{\mathcal{I}^b\}_{b=1}^B$ as input, the proposed scheme is designed to predict 6D object poses $\{\mathcal{T}^b\}_{b=1}^B$, with each pose $\mathcal{T} = [\bm{R}|\bm{t}]$ represented by a 3D rotation $\bm{R} \in SO(3)$ and a 3D translation $\bm{t} \in \mathbb{R}^3$. 
The whole network architecture is shown in Fig. \ref{fig:model_arch}, which consists of three main modules, including a \textbf{Feature Extractor}, a \textbf{Coarse Classifier} of discretized pose anchors, and a \textbf{Fine Regressor of Residual Poses} to the nearest anchor. 
 
More specifically, we employ the same feature extractor as \cite{labbe2020cosypose} to learn the pose-sensitive feature vectors $\{\bm{f}^b \in \mathbb{R}^C \}_{b=1}^B$ from each frame, which are then fed into the decoupled coarse classifier and fine regressor individually, whose output are combined together as final pose predictions. 
The former learns coarse poses via classification on the discretized pose anchors, while the latter learns residual poses to refine the coarse ones of pose anchors locally; both modules share the same input features, achieving superior robustness against inconsistent feature scaling. We will take a single image as an example to detail the two pose estimation modules shortly, and thus omit the superscript $b$ of the notations for simplicity in the following subsections.

\paragraph{Coarse Classification on Discretized Pose Anchors.} Given the pose-sensitive feature $\bm{f}$ of $\mathcal{I}$, the goal of this module is to globally make coarse predictions of $\bm{R}$ and $\bm{t}$ via classification on their pre-defined anchors, respectively. 
For the rotation $\bm{R}$, we generate $N_{\bm{R}}$ anchors that are uniformly distributed on the whole $SO(3)$ space as \cite{li2018unified}, which are denoted as $ \{\bm{R}_a^{1}, \cdots, \bm{R}_a^{N_{\bm{R}}}\}$. For the translation $\bm{t}$, we factorize it into three individual classification targets, including the two translation components $v_x$ and $v_y$ on the image coordinate system along X-axis and Y-axis, with the remaining component $z$ along Z-axis; for each classification target $t \in \{v_x, v_y, z\}$, we discretize the range of $[d_t^{min}, d_t^{max}]$ into $N_t$ bins uniformly, and use the bin centers $\{t_a^{1}, \cdots, t_a^{N_t}\}$ as the anchors of $t$.
We implement the classifier as four Multilayer Perceptrons (MLPs) with $N_{\bm{R}}, N_{v_x}, N_{v_y}, N_z$ output neurons, which are collectively denoted as the probabilities $\bm{S}_{\bm{R}}\in \mathbb{R}^{N_{\bm{R}}}$, $\bm{S}_{v_x}\in \mathbb{R}^{N_{v_x}}$, $\bm{S}_{v_y}\in \mathbb{R}^{N_{v_y}}$, and $\bm{S}_{z}\in \mathbb{R}^{N_{z}}$ of $\bm{R}$, $v_x$, $v_y$ and $z$, respectively. Denoting their indexes of maximal probabilities as $i_{\bm{R}}^{max}$, $i_{v_x}^{max}$, $i_{v_y}^{max}$ and $i_{z}^{max}$, the classifier finally gives out their coarse pose predictions as $\bm{R}_{cls} = \bm{R}^{i_{\bm{R}}^{max}}_{a}$, $v_{x, cls} = v_{x, a}^{i_{v_x}^{max}}$, $v_{y, cls} = v_{y, a}^{i_{v_y}^{max}}$ and $z_{cls} = z_{a}^{i_{z}^{max}}$.

\paragraph{Fine Regressor of Residual Poses.} This module shares the same input feature $\bm{f}$ as the coarse classifier to make the learning more robust to feature scale variations, and is implemented as four MLPs with $N_{\bm{R}} \times 6, N_{v_x}, N_{v_y}, N_z$ output neurons to regress the residuals of the pose anchors. 
We collectively denote the outputs as $\{\bm{R}_{reg,6D}^{i}\}_{i=1}^{N_{\bm{R}}}$, $\{v_{x,reg}^{i}\}_{i=1}^{N_{v_x}}$, $\{v_{y,reg}^{i}\}_{i=1}^{N_{v_y}}$, and $\{z_{reg}^{i}\}_{i=1}^{N_{z}}$; here we use the continuous 6D representations of rotation \cite{zhou2019continuity} as the regression target, which can be transformed into rotation matrices $\{\bm{R}_{reg}^{i}\}_{i=1}^{N_{\bm{R}}}$. 
According to probabilities of the classifier, the fine regressor refines the coarse predictions via the residuals $\bm{R}_{reg} = \bm{R}^{i_{\bm{R}}^{max}}_{reg}$, $v_{x, reg} = v_{x, reg}^{i_{v_x}^{max}}$, $v_{y, reg} = v_{y, reg}^{i_{v_y}^{max}}$, and $z_{reg} = z_{reg}^{i_{z}^{max}}$. 

Combining coarse anchor predictions and their residuals, our proposed network can generate the final object pose $\mathcal{T}=[\bm{R}|\bm{t}]$, with $\bm{t}=[x,y,z]$, as follows:
\begin{equation}
    \left\{\begin{aligned}
         \bm{R} &  = \bm{R}_{reg} \cdot \bm{R}_{cls} \\
        x & = (v_{x,cls} + v_{x, reg}) \cdot z / f_x \\
        y & = (v_{y,cls} + v_{y, reg}) \cdot z / f_y \\
        z & = z_{cls} + z_{reg}
       \end{aligned}\right.,
\end{equation}
where $f_x$ and $f_y$ are the focal lengths along X-axis and Y-axis, respectively.

\subsection{Manifold-Aware Objective}
\label{subsec:objective}

To train our network, we formulate the following manifold-aware objective $\mathcal{L}$ via combining a \textbf{coarse-to-fine pose decomposition loss} $\mathcal{L}_{pose}$ with a \textbf{cumulative target correlation regularization} $\mathcal{L}_{ctc}$:
\begin{equation}
    \mathcal{L} = \mathcal{L}_{pose} + \mathcal{L}_{ctc},
    \label{eqn:obj}
\end{equation}
where {$\mathcal{L}_{pose}$ favors for domain-invariant representations in 6D pose estimation across domains, while $\mathcal{L}_{ctc}$ enforces target manifolds into representation learning.}

\paragraph{Coarse-to-fine Pose Decomposition Loss.} $\mathcal{L}_{pose}$ consists of two loss terms $\mathcal{L}_{cls}$ and $\mathcal{L}_{reg}$ for the coarse classifier and the fine regressor, respectively, as follows:
\begin{equation}
    \mathcal{L}_{pose} = \frac{1}{B} \sum_{b=1}^B \mathcal{L}_{cls}^b + \mathcal{L}_{reg}^b.
    \label{eqn:obj_pose}
\end{equation}
For simplicity, we introduce $\mathcal{L}_{pose}$ on single input, and thus omit the batch index $b$ accordingly.

For the coarse classifier, given the ground truth pose $\tilde{\mathcal{T}} = [\tilde{\bm{R}}|\tilde{\bm{t}}]$, with $\tilde{\bm{t}}=[\tilde{x}, \tilde{y}, \tilde{z}]$ (and $\tilde{v}_x, \tilde{v}_y$), we first adopt a sparse scoring strategy to assign the labels for $\bm{S}_{\bm{R}}$, $\bm{S}_{v_x}$, $\bm{S}_{v_y}$ and $\bm{S}_z$, resulting in $\tilde{\bm{S}}_{\bm{R}}$, $\tilde{\bm{S}}_{v_x}$, $\tilde{\bm{S}}_{v_y}$ and $\tilde{\bm{S}}_z$, respectively, with each element $\tilde{s}_t^i$ $(t\in \{\bm{R},v_x,v_y,z\})$ assigned as follows:
\begin{equation}
    \tilde{s}_t^i=\left\{
    \begin{array}{lll}
    \theta_{t,1},     &      & {i \in \text{NN}_1(\tilde{t})}\\
    \theta_{t,2},     &      & {i \in \text{NN}_{k_t}(\tilde{t})\backslash \text{NN}_1(\tilde{t})}\\
    0,            &      & {Otherwise}
    \end{array} \right.,
    \label{eqn:score_assignment}
\end{equation}
where $\theta_{t,1}\gg\theta_{t,2}$, and $\theta_{t,1}+(k_t-1)\theta_{t,2}=1$. 
$\text{NN}_{k_t}(\tilde{t})$ denotes the set of indexes of the $k_t$ nearest anchors of $\tilde{t}$.\footnote{We use the geodesic distance \cite{gao2018occlusion} to measure the distance of two rotations $\bm{R}_1$ and  $\bm{R}_2$ as $\arccos(\frac{trace(\bm{R}_1\bm{R}_2^T)-1}{2})$, and use the difference value to measure that of two scalars.} With the assigned labels, we use the cross-entropy loss $\mathcal{H}$ on top of the classifier as follows:


\begin{equation}
    \mathcal{L}_{cls} = \sum_{t\in \{\bm{R},v_x,v_y,z\}}\mathcal{H}(\bm{S}_t, \tilde{\bm{S}}_t).
\end{equation}

For the fine regressor, we make individual predictions on each anchor of $t\in\{\bm{R},(v_x,v_y),z\}$ by combining the paired classification and regression results, and supervise the predictions of their top K nearest anchors of $\tilde{t}$ as follows:
\begin{equation}
    \label{eqn:loss_reg}
    \begin{aligned}
        \mathcal{L}_{reg} = &\sum_{i \in \text{NN}_{k_{\bm{R}}}(\tilde{\bm{R}})}\mathcal{D}(\mathcal{T}_{\bm{R}^i}, \tilde{\mathcal{T}}) + \sum_{i \in \text{NN}_{k_z}(\tilde{z})} \mathcal{D}(\mathcal{T}_{z^i}, \tilde{\mathcal{T}})\\
         &+ \sum_{i \in \text{NN}_{k_{v_xv_y}}(\tilde{v}_x \tilde{v}_y)}\mathcal{D}(\mathcal{T}_{v_x^i v_y^i}, \tilde{\mathcal{T}}), 
    \end{aligned}
\end{equation}
where $t^i$ denotes the prediction of the anchor $i$ of $t$, and $\mathcal{T}_{t^i}$ denotes the object pose computed by $t^i$ and other ground truths $\{\tilde{\bm{R}}, (\tilde{v}_x, \tilde{v}_y), \tilde{z}\} \backslash \tilde{t}$.  $\mathcal{D}(\cdot, \cdot)$ is the $L_1$ distance between the point sets transformed by two object poses from the same object point cloud $\mathcal{O}$, as follows:
\begin{equation}
    \label{eqn:l1_dis_point_set}
    \mathcal{D}(\mathcal{T}, \tilde{\mathcal{T}}) = \frac{1}{|\mathcal{O}|}\sum_{x\in \mathcal{O}}\|\mathcal{T} x - \tilde{\mathcal{T}} x\|_1.
\end{equation}
Following \cite{labbe2020cosypose}, we combine the supervision of $v_x$ and $v_y$ for convenience in (\ref{eqn:loss_reg}), and also employ the same strategy to handle object symmetries by finding the closest ground truth rotation to the predicted one.

\paragraph{Cumulative Target Correlation Regularization.} {
For regression tasks, continuous targets preserve latent cumulative dependency \cite{chen2013cumulative}.
When we discretize the continuously changing targets into discretized labels as classification, {the assumption of independence across targets is adopted, which is invalid in regressing continuous targets.}
As a result, each class cannot seek support from samples of correlated class, which can significantly reduce performance especially for sparse and imbalanced data distributions.
To better cope with this problem, we propose to regularize the features by an explicit relation in the regression target space.}

Given the pose-sensitive feature vectors $\{\bm{f}^b \in \mathbb{R}^C \}_{b=1}^B$ of a mini-batch inputs $\{\mathcal{I}^b\}_{b=1}^B$, we first build the feature correlation graph $\mathcal{G} \in \mathbb{R}^{B\times B}$ across the data batch via feature cosine similarities, with the element $g^{ij}$ indexed by $(i,j)$ computed as follows:
\begin{equation}
    g^{ij} = \frac{<\bm{f}^i, \bm{f}^j>}{||\bm{f}^i||_2 \cdot ||\bm{f}^j||_2},
\end{equation}
where $<\cdot,\cdot>$ denotes inner product. 
We then build the ground truth $\tilde{\mathcal{G}}$ based on a pre-computed correlation graph $\tilde{\mathcal{G}}_{0} \in \mathbb{R}^{N\times N}$ with $N$ pose classes; assuming the classes of $\mathcal{I}_i$ and $\mathcal{I}_j$ are $n_i$ and $n_j$, respectively, we assign the value of $\tilde{g}^{ij} \in \tilde{\mathcal{G}}$ as that of $\tilde{g}_0^{n_in_j}$.
Finally, the proposed target correlation regularizer can be simply written as the squared $L_2$ distance between $\mathcal{G}$ and $\tilde{\mathcal{G}}$:
\begin{equation}
    \mathcal{L}_{ctc} = \|\mathcal{G} - \tilde{\mathcal{G}} \|_2^2.
\end{equation}

There are multiple choices for building the pose-related correlation graph $\tilde{\mathcal{G}}_{0}$; here we introduce a simple but effective one, which utilizes the similarity of depth components of translations along Z-axis to initialize $\tilde{\mathcal{G}}_{0}$, with $N = N_z$. Specifically, for the anchors $\{z_a^{1}, \cdots, z_a^{N}\}$ of $z$, we map them linearly to the angles $\{\phi^{1}, \cdots, \phi^{N}\}$ as follows:
\begin{equation}
    \phi^{n} = \frac{z_a^n}{z^{max}-z^{min}} \cdot \frac{\pi}{2},
\end{equation}
and the element $\tilde{g}_0^{n_in_j}$ of $\tilde{\mathcal{G}}_{0}$ indexed by $(n_i, n_j)$ can be defined as the cosine of difference between the angles:
\begin{equation}
    \tilde{g}_0^{n_in_j} = \cos (|\phi^{n_i}-\phi^{n_j}|).
\end{equation}
When $z_a^{n_i}$ and $z_a^{n_j}$ are close, the difference of their corresponding angles is small, and thus the correlation value of $\tilde{g}_0^{n_in_j}$ will be large. The reason for choosing $z$ is that the learning of this component is very challenging in 6D pose estimation without depth information. Experimental results in Sec. \ref{subsec:ablation} also verify the effectiveness of our  regularization.

\subsection{Manifold-Aware Self-training}
\label{subsec:self-training}

To reduce the Sim2Real domain gap, we design a manifold-aware self-training scheme for unsupervisedly adapting the pose estimator, 
which adaptively incorporates our proposed manifold-aware training objective in (\ref{eqn:obj}) with Self-Paced Self-Training \cite{zou2018unsupervised} to select target samples in an easy-to-hard manner.
More specifically, we first train a teacher model $\mathcal{M}_T$ on the labeled synthetic data (source domain) as a pseudo-label annotator for the unlabeled real-world data (target domain), and select the training samples from the real data with pseudo labels for the learning of a student model $\mathcal{M}_S$. 
Both teacher and student models share the same networks introduced in Sec. \ref{subsec:network}, and are trained by solving the problems of $\min_{\mathcal{M}_T} \mathcal{L}$ and $\min_{\mathcal{M}_S} \mathcal{L}$, respectively. 

The core of sample selection on the target domain lies on the qualities of pseudo labels. For the tasks of visual classification, the categorical probabilities are usually used as the measurement of qualities, while for those of visual regression tasks, \eg, object pose estimation in this paper, direct usage of the typical mean square error (MSE) can be less effective due to lack of directional constraints for adaptation.
In geometric viewpoint, the surface of a super ball can have the same MSE distance to its origin, but the optimal regions of object surface for domain adaptation exist, which can be omitted by the MSE metric.
Owing to the decomposition of object pose estimation into coarse classification and fine regression in our MAST scheme, we can flexibly exploit the classification scores to indicate the qualities of pseudo labels, since the coarse classification points out the overall direction of pose estimation. 
In practice, we use the probabilities $\bm{S}_z$ as confidence scores because UDA on classification can perform more stably and robustly, and set a threshold $\tau$ to select the samples with scores larger than $\tau$ for training $\mathcal{M}_S$. 
Larger score indicates higher quality of the pseudo label. 
Following \cite{zou2018unsupervised}, the threshold $\tau$ is gradually decreased during training, realizing the learning in an easy-to-hard manner and making $\mathcal{M}_S$ generalized to harder target samples.

\section{Experiments}

\begin{table*}[htbp]
\centering
\resizebox{\linewidth}{!}{
\begin{tabular}{lcccccccccccccc}
\hlineB{3}
\multicolumn{1}{c|}{Method} &
  Ape &
  Bvise &
  Cam &
  Can &
  Cat &
  Drill &
  Duck &
  \textit{Eggbox} &
  \textit{Glue} &
  Holep &
  Iron &
  Lamp &
  \multicolumn{1}{c|}{Phone} &
  Mean \\ \hline
\multicolumn{1}{l}{} &\multicolumn{14}{c}{Data: syn (w/ GT)
} \\
\multicolumn{1}{l|}{AAE \cite{sundermeyer2020augmented}} &
  4.0 &
  20.9 &
  30.5 &
  35.9 &
  17.9 &
  24.0 &
  4.9 &
  81.0 &
  45.5 &
  17.6 &
  32.0 &
  60.5 &
  \multicolumn{1}{c|}{33.8} &
  31.4 \\

\multicolumn{1}{l|}{DSC-PoseNet \cite{yang2021dsc}} &
  23.4 &
  75.6 &
  11.7 &
  40.1 &
  26.7 &
  53.8 &
  14.0 &
  73.6 &
  26.7 &
  19.5 &
  56.2 &
  39.4 &
  \multicolumn{1}{c|}{20.0} &
  37.0 \\

\multicolumn{1}{l|}{MHP \cite{manhardt2019explaining}} &
  11.9 &
  66.2 &
  22.4 &
  59.8 &
  26.9 &
  44.6 &
  8.3 &
  55.7 &
  54.6 &
  15.5 &
  60.8 &
  - &
  \multicolumn{1}{c|}{34.4} &
  38.8 \\
\multicolumn{1}{l|}{DPOD \cite{zakharov2019dpod}} &
  35.1 &
  59.4 &
  15.5 &
  48.8 &
  28.1 &
  59.3 &
  25.6 &
  51.2 &
  34.6 &
  17.7 &
  84.7 &
  45.0 &
  \multicolumn{1}{c|}{20.9} &
  40.5 \\

\multicolumn{1}{l|}{SD-Pose \cite{li2021sd}} &
  54.0 &
  76.4 &
  50.2 &
  81.2 &
  71.0 &
  64.2 &
  54.0 &
  \textbf{93.9} &
  \textbf{92.6} &
  24.0 &
  77.0 &
  82.6 &
  \multicolumn{1}{c|}{53.7} &
  67.3 \\

\multicolumn{1}{l|}{$\text{Self6D++}$ \cite{wang2021occlusion}} &
  50.9 &
  \textbf{99.4} &
  \textbf{89.2} &
  97.2 &
  79.9 &
  \textbf{98.7} &
  24.6 &
  81.1 &
  81.2 &
  41.9 &
  \textbf{98.8} &
  \textbf{98.9} &
  \multicolumn{1}{c|}{64.3} &
  77.4 \\
 
 \multicolumn{1}{l|}{$\text{MAR}$ (ours)} &
  68.6 &
  97.4 &
  79.4 &
  \textbf{98.3} &
  87.1 &
  94.2 &
  \textbf{61.3} &
  82.0 &
  87.1 &
  \textbf{56.7} &
  94.3 &
  92.3 &
  \multicolumn{1}{c|}{68.8} &
  \textbf{82.1} \\ \hline

\multicolumn{1}{l}{} &
\multicolumn{14}{c}{Data: syn (w/ GT) + real (w/ GT)} \\
\multicolumn{1}{l|}{DPOD \cite{zakharov2019dpod}} &
  53.3 &
  95.2 &
  90.0 &
  94.1 &
  60.4 &
  97.4 &
  66.0 &
  99.6 &
  93.8 &
  64.9 &
  \textbf{99.8} &
  88.1 &
  \multicolumn{1}{c|}{71.4} &
  82.6 \\
\multicolumn{1}{l|}{DSC-PoseNet \cite{yang2021dsc}} &
  59.2 &
  98.1 &
  88.0 &
  92.1 &
  79.4 &
  94.5 &
  51.7 &
  98.5 &
  93.9 &
  78.4 &
  96.2 &
  96.3 &
  \multicolumn{1}{c|}{90.0} &
  85.9 \\
  
\multicolumn{1}{l|}{$\text{Self6D++}$ \cite{wang2021occlusion}} &
  \textbf{85.0} &
  99.8 &
  \textbf{96.5} &
  99.3 &
  93.0 &
  \textbf{100.0} &
  65.3 &
  \textbf{99.9} &
  98.1 &
  73.4 &
  86.9 &
  99.6 &
  \multicolumn{1}{c|}{86.3} &
  91.0 \\

\multicolumn{1}{l|}{MAR (ours)} &
  81.4 &
  \textbf{99.9} &
  90.7 &
  \textbf{99.6} &
  \textbf{94.6} &
  98.1 &
  \textbf{85.5} &
  97.6 &
  \textbf{98.5} &
  \textbf{89.2} &
  97.1 &
  \textbf{99.7} &
  \multicolumn{1}{c|}{\textbf{96.0}} &
  \textbf{94.5} \\ \hline
 
\multicolumn{1}{l}{} &
\multicolumn{14}{c}{Data: syn (w/ GT) + real (w/o GT) } \\
\multicolumn{1}{l|}{Self6D-RGB \cite{wang2020self6d}} &
  0.0 &
  10.1 &
  3.1 &
  0.0 &
  0.0 &
  7.5 &
  0.1 &
  33.0 &
  0.2 &
  0.0 &
  5.9 &
  20.7 &
  \multicolumn{1}{c|}{2.4} &
  6.4 \\
\multicolumn{1}{l|}{DSC-PoseNet \cite{yang2021dsc}} &
  35.9 &
  83.1 &
  51.5 &
  61.0 &
  45.0 &
  68.0 &
  27.6 &
  89.2 &
  52.5 &
  26.4 &
  56.3 &
  68.7 &
  \multicolumn{1}{c|}{46.3} &
  54.7 \\
\multicolumn{1}{l|}{Zhang \etal \cite{zhang2021keypoint}} &
  -- &
  -- &
  -- &
  -- &
  -- &
  -- &
  -- &
  -- &
  -- &
  -- &
  -- &
  -- &
  \multicolumn{1}{c|}{-} &
  60.4 \\
\multicolumn{1}{l|}{Sock \etal \cite{sock2020introducing}} &
  37.6 &
  78.6 &
  65.5 &
  65.6 &
  52.5 &
  48.8 &
  35.1 &
  89.2 &
  64.5 &
  41.5 &
  80.9 &
  70.7 &
  \multicolumn{1}{c|}{60.5} &
  60.6 \\

\multicolumn{1}{l|}{Self6D++ \cite{wang2021occlusion}} &
  \textbf{76.0} &
  91.6 &
  \textbf{97.1} &
  \textbf{99.8} &
  85.6 &
  \textbf{98.8} &
  56.5 &
  91.0 &
  92.2 &
  35.4 &
  \textbf{99.5} &
  97.4 &
  \multicolumn{1}{c|}{\textbf{91.8}} &
  85.6 \\ 
  
\multicolumn{1}{l|}{$\text{MAST}$ (ours)} &
  73.5 &
  \textbf{97.2} &
  80.8 &
  98.6 &
  \textbf{89.1} &
  93.9 &
  \textbf{66.9} &
  \textbf{95.3} &
  \textbf{95.4} &
  \textbf{69.8} &
  95.5 &
  \textbf{98.6} &
  \multicolumn{1}{c|}{79.1} &
  \textbf{87.2} \\ \hlineB{3}

\end{tabular}
}
  \caption{Comparative evaluation on the LineMOD dataset \wrt the Average Recall (\%) of ADD(-S). Symmetric object classes are in italic. `MAR' (manifold-aware regression) denotes our method without self-training.}
  \label{tab:compare_LM}
\end{table*}

\begin{table*}[ht]
\centering
\resizebox{\linewidth}{!}{
\begin{tabular}{c|cccccccc|c|ccc|c}
\hlineB{3}
\multirow{2}{*}{Method}
&
\multicolumn{9}{c|}{Occluded LineMOD} & \multicolumn{4}{c}{HomebrewedDB}
\\ 
\cline{2-14}
& Ape  & Can  & Cat  & Drill & Duck & \textit{Eggbox} & \textit{Glue} & \multicolumn{1}{c|}{Holep} & Mean & Bvise & Drill & \multicolumn{1}{c|}{Phone} & Mean \\ \hline
\multicolumn{1}{l}{} &
\multicolumn{13}{c}{Data: syn (w/ GT)} 
                                                                                                          
\\
\multicolumn{1}{l|}{DPOD \cite{zakharov2019dpod}}            & 2.3  & 4.0  & 1.2  & 7.2  & 10.5 & 4.4  & 12.9 & \multicolumn{1}{c|}{7.5}  & 6.3 & 52.9  & 37.8  & \multicolumn{1}{c|}{7.3}   & 32.7 \\
\multicolumn{1}{l|}{CDPN \cite{li2019cdpn}}                  & 20.0 & 15.1 & 16.4 & 22.2 & 5.0  & 36.1 & 27.9 & \multicolumn{1}{c|}{24.0} & 20.8 & -- & -- & -- & --\\
\multicolumn{1}{l|}{SD-Pose \cite{li2021sd}}                 & 21.5 & 56.7 & 17.0 & 44.4 & 27.6 & 42.8 & 45.2 & \multicolumn{1}{c|}{21.6} & 34.6 & -- & -- & -- & -- \\

\multicolumn{1}{l|}{SSD6D+Ref. \cite{manhardt2018deep}}                       
&-- &-- &-- &-- &-- &-- &-- &-- &-- & 82.0  & 22.9  & \multicolumn{1}{c|}{24.9}  & 43.3 \\                                                
                                                            
\multicolumn{1}{l|}{$\text{Self6D++}$ \cite{wang2021occlusion}}  
                                 & 44.0 & \textbf{83.9} & \textbf{49.1} & \textbf{88.5} & 15.0 & \textbf{33.9} & \textbf{75.0} & \multicolumn{1}{c|}{34.0} & 52.9 & 7.1   & 2.2   & \multicolumn{1}{c|}{0.1}   & 3.1 \\ 
                                                            
\multicolumn{1}{l|}{MAR (ours)}  & \textbf{44.9} & 78.4 & 40.3 & 73.5 & \textbf{47.9} & 26.9 & 72.1 & \multicolumn{1}{c|}{\textbf{58.0}} & \textbf{55.3} & \textbf{92.6}  & \textbf{91.5}  & \multicolumn{1}{c|}{\textbf{80.0}}  & \textbf{88.0}\\
\hline

\multicolumn{1}{l}{} &                                                      
\multicolumn{13}{c}{Data: syn (w/ GT) + real (w/o GT)}  \\
\multicolumn{1}{l|}{DSC-PoseNet \cite{yang2021dsc}}         & 13.9 & 15.1 & 19.4 & 40.5 & 6.9  & 38.9 & 24.0 & \multicolumn{1}{c|}{16.3} & 21.9 & 72.9  & 40.6  & \multicolumn{1}{c|}{18.5}  & 44.0 \\
\multicolumn{1}{l|}{Sock \etal \cite{sock2020introducing}}   & 12.0 & 27.5 & 12.0 & 20.5 & 23.0 & 25.1 & 27.0 & \multicolumn{1}{c|}{35.0} & 22.8 & 57.3  & 46.6  & \multicolumn{1}{c|}{41.5}  & 52.0 \\
\multicolumn{1}{l|}{Zhang \etal \cite{zhang2021keypoint}}        & --    & --    & --    & --    & --    & --    & --    & \multicolumn{1}{c|}{--}    & 33.7 & --     & --     & \multicolumn{1}{c|}{--}     & 63.8 \\
\multicolumn{1}{l|}{Self6D++ \cite{wang2021occlusion}}       & \textbf{57.7} & \textbf{95.0} & \textbf{52.6} & \textbf{90.5} & 26.7 & 45.0 & \textbf{87.1} & \multicolumn{1}{c|}{23.5} & 59.8 & 56.1  & \textbf{97.7}  & \multicolumn{1}{c|}{\textbf{85.1}}  & 79.6 \\ 
\multicolumn{1}{l|}{MAST (ours)}                            & 47.6 & 82.9 & 45.4 & 75.0 & \textbf{53.7} & \textbf{48.2} & 75.3 & \multicolumn{1}{c|}{\textbf{63.0}} & \textbf{61.4} & \textbf{93.8}  & 91.5  & \multicolumn{1}{c|}{81.8}  & \textbf{89.0} \\ \hlineB{3}
\end{tabular}
}
  \caption{Comparative evaluation on the Occluded LineMOD and HomebrewedDB datasets \wrt the Average Recall (\%) of the ADD(-S). Symmetric object classes are in italic. `MAR' (manifold-aware regression) denotes our method without self-training.}
  \label{tab:compare_LMO-HB}
\end{table*}

\paragraph{Datasets and Settings.} The LineMOD dataset \cite{hinterstoisser2012model} provides individual videos of 13 texture-less objects, which are recorded in cluttered scenes with challenging lighting variations. For each object, we follow \cite{brachmann2014learning} to use randomly sampled $15\%$ of the sequence as the real-world training data of the target domain, and the remaining images are set aside for testing. 
The Occluded LineMOD dataset \cite{brachmann2014learning} is a subset of the LineMOD with 8 different objects, which is formed by the images with severe object occlusions and self-occlusions. We follow \cite{wang2021occlusion} to split the training and test sets.
The HomebrewedDB dataset \cite{kaskman2019homebreweddb} provides newly captured test images of three objects in the LineMOD, including bvise, driller and phone. Following the Self-6D \cite{wang2020self6d}, the second sequence of HomebrewedDB is used to test our models which are trained on the LineMOD, to evaluate the robustness of our method on different variations, \eg, scene layouts and camera intrinsics.
In the experiments, the above three real-world datasets are considered as the target domains, all of which share the same synthetic source domain. We employ the publicly available synthetic data provided by BOP challenge \cite{hodan2020bop} as the source data, which contains 50k images generated by physically-based rendering (PBR) \cite{hodavn2019photorealistic}.

\paragraph{Evaluation Metrics.}
Following \cite{wang2020self6d}, we employ the Average Distance of model points (ADD) \cite{hinterstoisser2012model} as the evaluation metric of the 6D poses for asymmetric objects, which measures the average deviation of the model point set $\mathcal{O}$ transformed by the estimated pose $\mathcal{T} = [\bm{R}|\bm{t}]$ and that transformed by the ground-truth pose $\tilde{\mathcal{T}} = [\tilde{\bm{R}}|\tilde{\bm{t}}]$:
\begin{equation}
    \label{eqn:add}
    \mathcal{D}_{\text{ADD}}(\mathcal{T}, \tilde{\mathcal{T}}) = \frac{1}{|\mathcal{O}|} \sum_{\bm{x} \in \mathcal{O}} \|(\bm{R}\bm{x} + \bm{t}) - (\tilde{\bm{R}}\bm{x} + \tilde{\bm{t}}) \|_2.
\end{equation}
For symmetric objects, we employ the metric of Average Distance of the closest points (ADD-S) \cite{hodavn2016evaluation}:
\begin{equation}
    \label{eqn:adds}
    \mathcal{D}_{\text{ADD-S}}(\mathcal{T}, \tilde{\mathcal{T}}) = \frac{1}{|\mathcal{O}|} \sum_{\bm{x}_1 \in \mathcal{O}} \min_{\bm{x}_2 \in \mathcal{O}} \|(\bm{R}\bm{x}_1 + \bm{t}) - (\tilde{\bm{R}}\bm{x}_2 + \tilde{\bm{t}}) \|_2.
\end{equation}
Combining (\ref{eqn:add}) and (\ref{eqn:adds}), we report the Average Recall ($\%$) of ADD(-S) less than $10\%$ of the object's diameter on all the three datasets.

\paragraph{Implementation Details.}
For object detection, we use Mask R-CNN \cite{he2017mask} trained purely on synthetic PBR images to generate the object bounding boxes for the target real data. 
For pose estimation, we set the numbers of anchors as $N_{\bm{R}}=60, N_{v_x}=N_{v_y}=20, N_z=40$, and set the ranges of $v_x$, $v_y$ and $z$ as $[d_{v_x}^{min}, d_{v_x}^{max}]=[d_{v_y}^{min}, d_{v_y}^{max}]=[-200, 200]$, and $[d_{z}^{min}, d_{z}^{max}]=[0.0, 2.0]$, respectively. 
To train our network, we choose $\theta^{\bm{R}}_1=0.7, \theta^{\bm{R}}_2=0.1$ and $k_{\bm{R}}=4$ for rotation in (\ref{eqn:score_assignment}), and also set $\theta^{v_x}_1 = \theta^{v_y}_1 = \theta^{z}_1=0.55, \theta^{v_x}_2 = \theta^{v_y}_2 = \theta^{z}_2=0.075$, and $k_{v_x}=k_{v_y}=k_z=7$ for translation. 
Following the popular setting \cite{wang2021occlusion}, we train individual networks for all the objects with the Adam optimizer \cite{kingma2014adam}. The teacher model $\mathcal{M}_T$ is firstly pre-trained on the synthetic images of all objects, and then fine-tuned on the single object, while the parameters of the student model $\mathcal{M}_S$ is initialized as those of $\mathcal{M}_T$; their initial learning rates are $3 \times 10 ^{-4}$ and $3 \times 10 ^{-5}$, respectively.  
The training batch size is set as $B=32$. We also include the same data augmentation as \cite{labbe2020cosypose} during training.

\subsection{Comparative Evaluation}
We compare our method with the existing ones on three benchmarks for 6D object pose estimation with RGB images.  

On the LineMOD, we conduct experiments under three settings of training data, including 1) labeled synthetic data, 2) labeled synthetic and real data, and 3) labeled synthetic data and unlabeled real data. 
Results of the first two settings are the lower and upper bounds of that of the last setting.
We report qualitative results of comparative methods in Table \ref{tab:compare_LM}, where our method outperforms its competitors by large margins under all the settings, \eg, with the respective improvements of $4.7\%$, $3.5\%$ and $1.6\%$ over the state-of-the-art Self6D++ \cite{wang2021occlusion}. 
On the Occluded LineMOD and the HomebrewedDB, results are shown in Table \ref{tab:compare_LMO-HB}, where our method consistently performs better than the existing ones on both datasets, demonstrating superior robustness of our method against occlusion and the generalization to new scenes and cameras.

\subsection{Ablation Studies and Analyses}
\label{subsec:ablation}

\paragraph{Effects of Decomposing into Coarse Classification and Fine Regression.} We decompose the problem of UDA on estimating object poses into a coarse classification on discretized anchors and a residual regression. 
As shown in Table \ref{tab:ablation_LM-LMO}, for the models trained purely on synthetic data, the design of pose decomposition realizes $4.0\%$ and $5.7\%$ improvements on the LineMOD and the Occluded LineMOD, respectively, compared to direct regression of object poses, {since the global classification eases the difficulty in learning along with feature-scaling robustness, and the local regression achieves pose refinement.}

\begin{table}[htbp]
\centering
\resizebox{\columnwidth}{!}{
  \begin{tabular}{c|c|c|c|c}
    \hlineB{3}
    \multirow{2}{*}{Pose Estimator} & \multirow{2}{*}{$\mathcal{L}_{ctc}$} & \multirow{2}{*}{Method of UDA} & \multicolumn{2}{c}{Dataset} \\
    \cline{4-5}
     &  &   & LM & LMO \\\hline
    \multicolumn{5}{c}{Data: syn (w/ GT)}\\
    Reg.  &        $\times$       & -            & 75.3          & 44.0    \\
    Cls. + Reg.  &   $\times$     & -        & 79.3          & 49.7    \\
    Cls. + Reg.  & \checkmark     & -       & \textbf{82.1} & \textbf{55.3}   \\ \hline
    \multicolumn{5}{c}{Data: syn (w/ GT) + real (w/o GT)}\\
    Cls. + Reg.  & \checkmark   & RSD \cite{chen2021representation}   & 83.9     & 55.0  \\
    Cls. + Reg.  &   $\times$   & Self-Training                            & 85.6                          & 60.1           \\
    Cls. + Reg.  & \checkmark   & Self-Training                          & \textbf{87.2}                 & \textbf{61.4} \\\hlineB{3}
\end{tabular}
}
  \caption{Ablation studies on LineMOD (LM) and Occluded LineMOD (LMO) datasets \wrt the Average Recall (\%) of ADD(-S). `Reg.' denotes direct regression of object poses, while `Cls. + Reg.' denotes the combined use of coarse classification and fine regressor for pose estimation.}
  \label{tab:ablation_LM-LMO}
\end{table}

\paragraph{Effects of Cumulative Target Correlation Regularization.} As shown in Table \ref{tab:ablation_LM-LMO}, $\mathcal{L}_{ctc}$ consistently improves the results under different settings across different datasets, \eg, $5.6\%$ improvement on the Occluded LineMOD for the model trained on synthetic data, which demonstrates the effectiveness of $\mathcal{L}_{ctc}$ on mining latent correlation across regression targets. 
We also visualize the feature distribution of an example via the t-SNE \cite{van2008visualizing} in Fig. \ref{fig:intro}, where, with $\mathcal{L}_{ctc}$, features assigned to different pose anchors preserve smooth and continuously changing nature of regression targets in the feature space.

\paragraph{Effects of Manifold-Aware Self-Training on Coarse Classification.} The self-training schemes have been verified their effectiveness on reducing the Sim2Real domain gap by incorporating the unlabeled real data into training via pseudo label generation and training sample selection.
Taking our network with $\mathcal{L}_{ctc}$ as example, the results are improved from $55.3\%$ to $61.4\%$ on the Occluded LineMOD via self-training. 
Compared to the RSD \cite{chen2021representation} designed for the problem of UDA on regression, our MAST scheme can significantly beat the competing RSD (see results in Table \ref{tab:ablation_LM-LMO}), where the only difference lies in replacing self-training on coarse classification with RSD on whole regression.
Such an observation can again confirm the superiority of the proposed MAST scheme, consistently outperforming the state-of-the-art UDA on regression.

\paragraph{On More Visual Regression Tasks.}
We conduct more experiments on the dSprites dataset \cite{matthey2017dsprites} for assessing UDAVR performance.
For simplicity, the problem aims to regress the "scale" variable of a shape from images. 
Using the same backbone as RSD, under the UDA setting from the scream (S) domain to the noisy (N) domain, our MAST can achieve 0.024 in terms of mean absolute error, while the RSD only obtains 0.043.

\paragraph{Run-time analysis.}
On a server with NVIDIA GeForce RTX 3090 GPU, given a 640 $\times$ 480 image, the run-time of our network is up to 5.8 ms/object including object detection and pose estimation when using Mask R-CNN \cite{he2017mask} as detector.
Pose estimation takes around 5 ms/object.

\paragraph{Details of output pose.}
We employ a render-and-compare style pose refinement process as \cite{labbe2020cosypose} to get final object pose.
An initial guess pose $[\bm{R}_{init}, x_{init}, y_{init}, z_{init}]$ is calculated from bounding box and object CAD model using the same strategy as \cite{labbe2020cosypose}. 
Given the network output $[\bm{R}, x, y, z]$, the estimated object pose $[\bm{R}_{obj}, x_{obj}, y_{obj}, z_{obj}]$ can be calculated by:
\begin{equation}
    \left\{\begin{aligned}
         \bm{R}_{obj} &= \bm{R} \cdot \bm{R}_{init} \\
        x_{obj} &= x + x_{init} \\
        y_{obj} &= y + y_{init} \\
        z_{obj} &= z \cdot z_{init}
       \end{aligned}\right.,
\end{equation}

\begin{figure}[t]
    \centering
    \includegraphics[width=0.95\columnwidth]{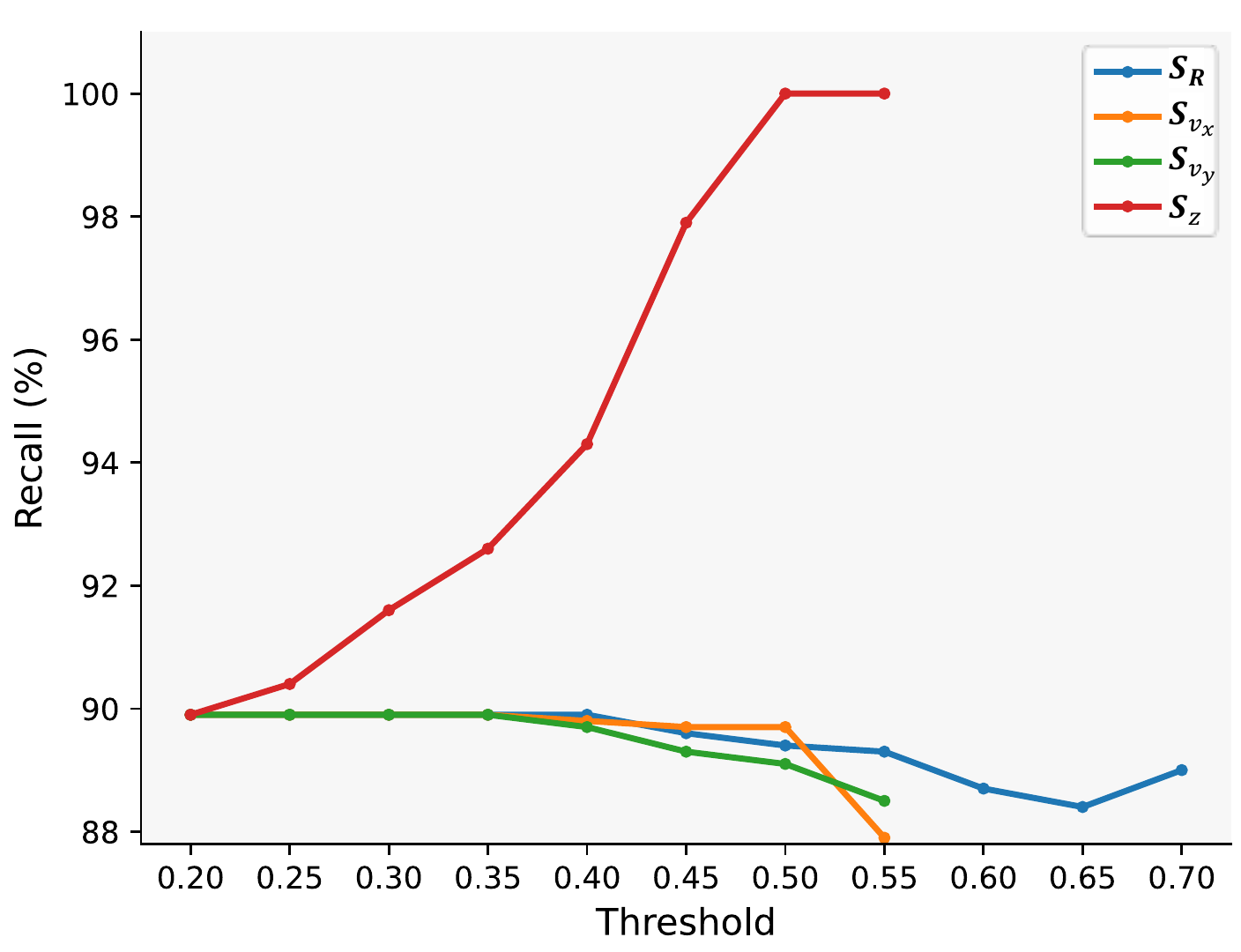}
    \caption{The effect of selecting samples via $\bm{S}_{\bm{R}}$,  $\bm{S}_{v_x}$,  $\bm{S}_{v_y}$,  $\bm{S}_z$ confidence w.r.t. the recall of the ADD(-S) among selected samples on driller object of LineMOD dataset training set.}
    \label{fig:thre_ar}
\end{figure}

\paragraph{On selecting samples with pseudo pose labels.}
We choose the probability $\bm{S}_z$ as confidence scores in practice, Fig. \ref{fig:thre_ar} shows the average recall of selected samples with pseudo pose labels via  $\bm{S}_{\bm{R}}$,  $\bm{S}_{v_x}$,  $\bm{S}_{v_y}$,  $\bm{S}_z$, which tells that as the confidence threshold becomes larger, only red line ($\bm{S}_z$) grows in terms of the average recall while others remain unchanged or decreased.

\section{Conclusion}
This paper proposes a novel and generic manifold-aware self-training scheme for UDA on regression, which is applied to the challenging 6D pose estimation of object instances. 
We address the UDAVR problem via decomposing it into coarse classification and fine regression, together with a cumulative target correlation regularization.
Experiment results on three popular benchmarks can verify the effectiveness of our MAST scheme, outperforming the state-of-the-art methods with significant margins.
{It is worth pointing out that our MAST scheme can readily be applied to any UDA regression tasks, as the UDA on coarse classification making our method robust against feature scaling while maintaining latent cumulative correlation underlying in regression target space.}

\section*{Acknowledgments}
This work is supported in part by the National Natural Science Foundation of China (Grant No.: 61902131), the Guangdong Youth Talent Program (Grant No.: 2019QN01X246), the Guangdong Basic and Applied Basic Research Foundation (Grant No.: 2022A1515011549), the Program for Guangdong Introducing Innovative and Enterpreneurial Teams (Grant No.: 2017ZT07X183), and the Guangdong Provincial Key Laboratory of Human Digital Twin (Grant No.: 2022B1212010004).

\bibliographystyle{named}
\bibliography{ijcai23}

\end{document}